\documentclass[sigconf,nonacm]{acmart}
\usepackage{algorithm}
\usepackage{algorithmic}
\usepackage{booktabs}

\usepackage{amsmath}
\usepackage{threeparttable}
\usepackage{amsfonts}
\usepackage{xcolor}

\usepackage{amssymb}
\usepackage{multirow}
\usepackage{makecell}
\usepackage{dsfont}
\usepackage{xspace}
\allowdisplaybreaks[3]

\title{COVER: Codec-Robust Video Watermarking with Generative Video Priors}

\author{Yuxin Cao}
\authornote{These authors contributed equally.}
\affiliation{%
  \institution{National University of Singapore}
  \country{Singapore}
}
\author{Hao Yang}
\authornotemark[1]
\affiliation{%
  \institution{Beijing University of Post and Telecommunications}
  \country{China}
}
\author{Ziqi Ding}
\affiliation{%
  \institution{University of New South Wales}
  \country{Australia}
}
\author{Jie Hao}
\affiliation{%
  \institution{Beijing University of Post and Telecommunications}
  \country{China}
}
\author{Wei Song}
\affiliation{%
  \institution{University of New South Wales}
  \country{Australia}
}

\begin{document}

\begin{abstract}
Video watermarking underpins copyright protection and provenance for generated media, yet almost every video is compressed by a codec before it is stored or shared. A codec discards precisely the perceptually redundant components that most watermarks rely on, so the payload is often lost even when the marked video looked flawless beforehand. Existing methods leave this path open, since they treat compression as one entry in a generic list of distortions, while a real codec is not differentiable and cannot enter gradient-based training. We present COVER, the first learned video watermark built around codec compression as its design target, which survives that compression by embedding the payload in the latent space of a frozen generative video autoencoder and recovering it by re-encoding the received video into that same latent space. To make codec robustness trainable, we build a differentiable codec surrogate bank that simulates the dominant degradation modes of practical compression, and we train the embedder and the latent decoder through three shared recovery paths under a fidelity objective that constrains the residual in the pixel and frequency domains. Across four codecs at 12 settings, COVER attains 93.72\% average bit accuracy, ranks first on 11 of the 12, improves the strongest prior method by 2.68 points, and lifts the worst operating point from 68.90\% to 73.72\% while each marked video stays visually close to the source clip that produced it.
\end{abstract}

\maketitle

\section{Introduction}

Generative video models now synthesize long, high resolution clips with coherent motion and semantics~\citep{cogvideox2024,latentdiffusion2022}, and video platforms increasingly rely on automated systems to produce and moderate such content at scale. This progress raises pressing questions about the ownership and authenticity of synthetic media~\citep{provenancehash2026}, and it makes reliable video watermarking a core requirement for identifying and tracing generated content~\citep{videoseal2024,videoshield2025,sigmark2026}. A watermark is useful only when its payload survives the processing a video meets on its way to a viewer~\citep{waves2024}, so robustness to that everyday processing, rather than imperceptibility alone, now decides whether a deployed watermark is worth anything at all.

Learning-based watermarking has largely replaced handcrafted embedding rules. Early neural schemes cast embedding and extraction as an end-to-end problem trained against differentiable distortion layers~\citep{hidden2018,stegastamp2020}, and later work steadily strengthened their robustness~\citep{romark2019,hinet2021,rosteals2023}. With the rise of generative models, recent methods place the watermark inside the generation process itself~\citep{stablesignature2023,treering2023,gaussianshading2024}, and a growing line of work extends the idea to video by modeling temporal structure and generation-aware embedding~\citep{videoseal2024,videoshield2025}. These systems show that a learned watermark can stay imperceptible under many distortions, yet the distortions that they train and report against remain the image-level operations inherited from work on still images.

\begin{figure*}[t]
    \centering
    \includegraphics[width=0.9\textwidth]{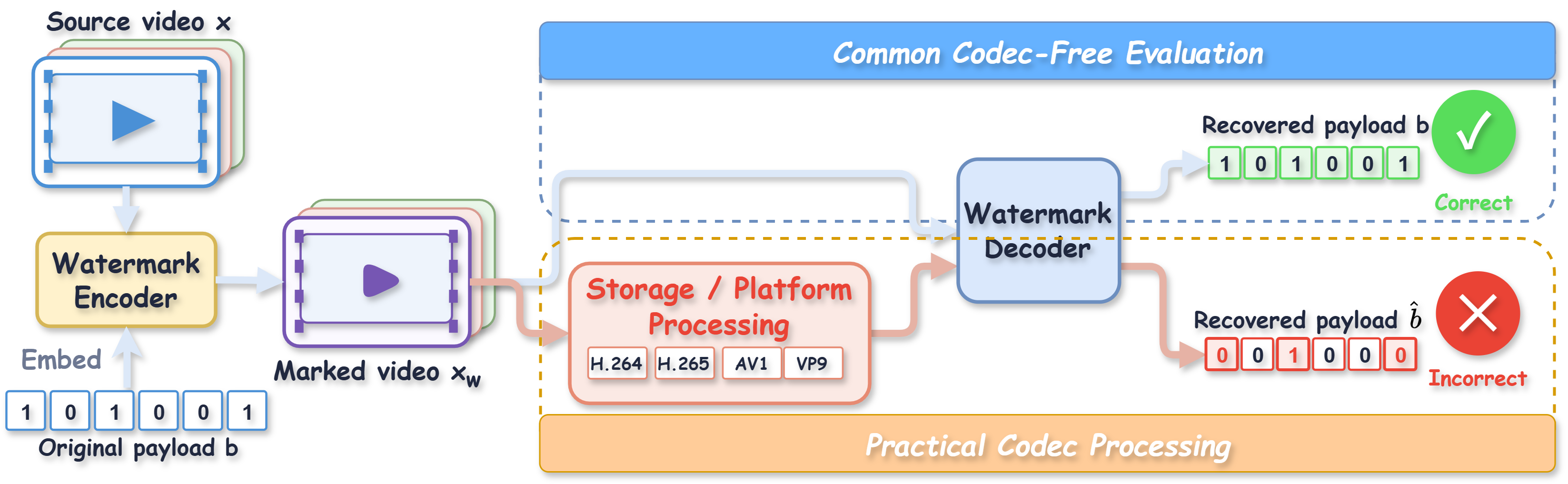}
    \caption{The gap this paper targets. A watermark is usually verified on the marked video itself, whereas a real video is compressed before it is stored or shared, so a payload that reads back perfectly on the first path can be lost entirely on the second, which is the path that a real viewer sees.}
    \label{fig:scenario}
\end{figure*}

A gap therefore remains on the one step that almost every real video must pass through, namely codec compression, as Figure~\ref{fig:scenario} illustrates. When a marked video is saved or shared, it is re-encoded by a codec that discards the fine spatial and temporal detail a viewer is least likely to notice, which is exactly where many watermarks place their energy, so the payload weakens or disappears even though the marked video looked flawless beforehand. Two obstacles have kept this failure open, and both of them come from the codec itself. A modern codec degrades a clip by composing several mechanisms at once and by adapting each of them to the content and the target bit rate~\citep{h264overview2003,hevcoverview2012}, so no single fixed distortion of the kind used in image watermarking stands in for it. A real codec is also not differentiable, so it enters gradient-based training only through a proxy. Prior video watermarks list compression among the distortions they train on, and none makes surviving a real codec the objective the whole design answers to.

\begin{figure*}[t]
    \centering
    \includegraphics[width=0.98\linewidth]{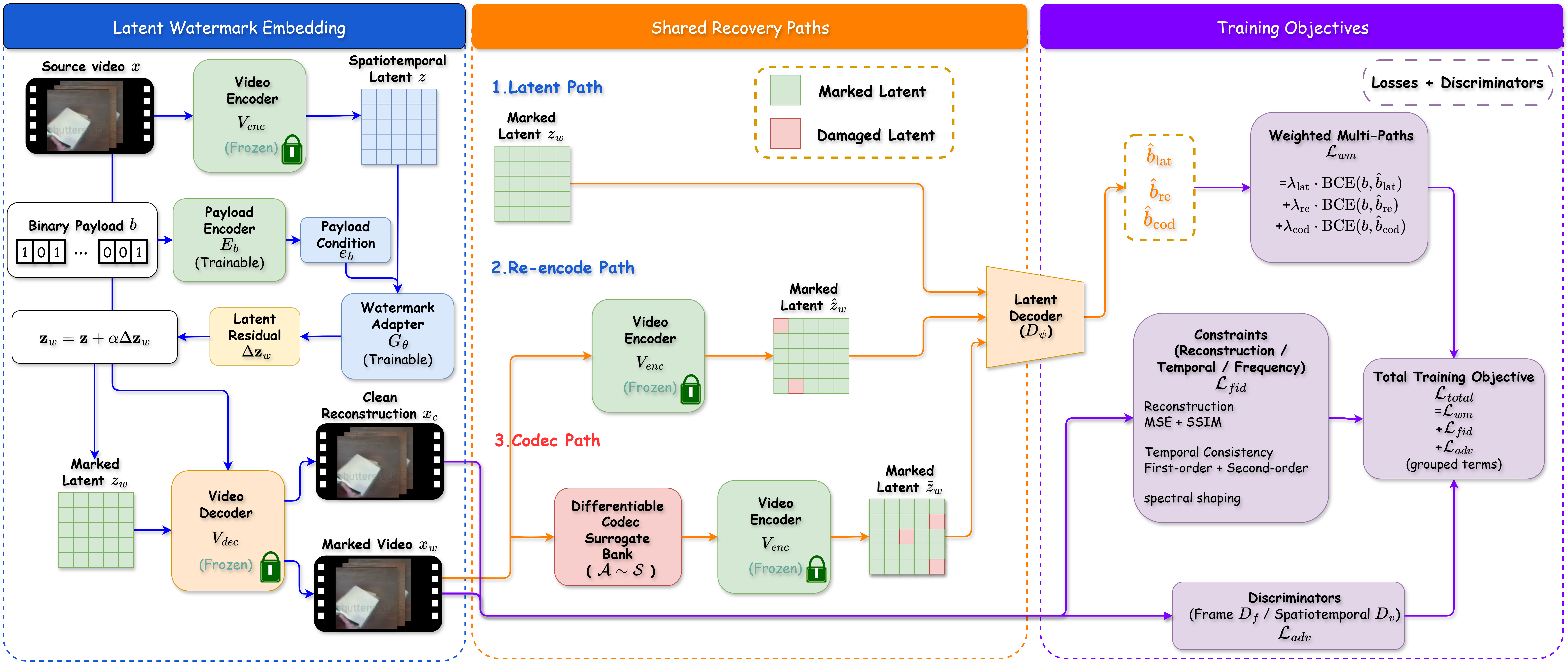}
    \caption{Overview of COVER. The payload is written into the latent of a frozen video autoencoder as a residual, and one shared latent decoder reads it back on three paths that expose it to the autoencoder round trip and to a differentiable codec surrogate.}
    \label{fig:framework}
\end{figure*}

We take that position with COVER (\underline{C}odec-R\underline{o}bust \underline{V}id\underline{e}o Wate\underline{r}marking), which moves the watermark away from fragile pixel detail and into the latent space of a frozen video autoencoder~\citep{cogvideox2024}. A lightweight adapter writes the payload into the latent as a residual, and detection re-encodes the received video into that same space so a latent decoder can read the payload back, as Figure~\ref{fig:framework} shows. The design follows the way a codec spends its bit budget. Compression keeps the low and mid frequency structure that shapes what a viewer sees and drops the detail a viewer rarely notices, and a generative video prior organizes a clip around that same perceptually important structure, so embedding in the latent places the payload where the codec has the least reason to remove it and detection never has to undo quantization in the pixel domain, where the discarded detail is gone for good.

COVER makes this trainable with a differentiable codec surrogate bank that simulates the degradation of modern codecs, so the embedder and the latent decoder learn against realistic compression instead of a single image distortion. One shared latent decoder reads the payload from the marked latent, from its re-encoding, and from its re-encoding after the surrogate, and the objective couples payload recovery with fidelity terms in the pixel and frequency domains. We then evaluate against real compression rather than against the surrogate, on 500 videos passed through four codec families at 12 strengths, where COVER leads every family and widens its advantage exactly at the settings where compression is most aggressive and the payload is at its hardest to read back.

Our main contributions are summarized as follows.
\begin{itemize}
    \item We make codec compression the design target of a video watermark rather than one distortion among many, and realize it in COVER, which embeds a payload in a frozen video latent and reads it back from the re-encoded video.
    \item We build a differentiable codec surrogate bank that models the dominant degradation modes of a codec, which turns robustness to real compression into a trainable objective across three recovery paths that share one latent decoder.
    \item We evaluate on 500 videos under four real codecs at 12 settings, where COVER ranks first on 11 of them and improves the strongest prior method by 2.68 points, lifting the worst of the 12 settings from 68.90\% to 73.72\%.
\end{itemize}

\section{Related Work}
\label{sec:related_work}

\paragraph{Deep Image and Generative Watermarking.}
Learning-based methods replaced handcrafted embedding rules by training an encoder and a decoder jointly against a differentiable distortion layer, an idea introduced by HiDDeN and StegaStamp~\citep{hidden2018,stegastamp2020} and steadily strengthened since~\citep{distortionagnostic2020,mbrs2021,hinet2021,rosteals2023,watermarkanything2025,trustmark2025}. Diffusion models then moved the watermark from a post-processing step into generation itself, rooting it in the latent decoder or in the initial noise~\citep{stablesignature2023,treering2023,gaussianshading2024,wmadapter2025}, and later work showed that a signal placed on a generative prior survives heavy image editing~\citep{lawa2024,vine2025,gaussmarker2025}. This line establishes the generative prior as a stable place to hide a payload, yet it targets images throughout and models neither temporal structure nor the degradation that a real codec inflicts.

\paragraph{Video Watermarking.}
Video asks the payload to survive spatial content and temporal change. Early systems carried image techniques across frames through attention or temporal residuals~\citep{rivagan2019,videosteg2019}, whereas recent methods build the watermark into modern video pipelines. Video Seal trains an efficient embedder and extractor for general videos~\citep{videoseal2024}, and LVMark writes the payload into the latent of a video diffusion model~\citep{lvmark2024}. VideoShield and VideoMark instead plant the payload in the noise a generator starts from and read it back by inverting generation~\citep{videoshield2025,videomark2025}, so they hold no trainable watermark parameter and can mark only the clips they synthesize, and VideoSignature follows the same route with an implicit signature~\citep{videosignature2025}. Newer systems continue to strengthen watermarking inside video generation~\citep{sigmark2026,skeda2026,unifywm2026}. All treat compression as one distortion among many and read the payload from processed frames rather than a re-encoded latent, and this leaves the codec channel that we target essentially open right across the line of work that we have just described.

\paragraph{Robustness against Compression.}
Lossy compression is the hardest distortion for a watermark, since it removes the perceptually redundant signal a watermark relies on to stay hidden. For images, robustness to a codec is usually learned by approximating it with a differentiable proxy or by mixing real and simulated compression into training~\citep{jpegstandard1992,distortionagnostic2020,mbrs2021,flowwatermark2023}. A video codec is far harder to imitate, since it couples spatial transform coding with motion compensation across frames~\citep{h264overview2003,hevcoverview2012,deepcoder2017,dvc2019,compressionreview2020}, so a single fixed proxy cannot represent it while the codec itself provides no gradient. Classical video watermarking answered this by writing the payload into the compressed bitstream of a specific standard~\citep{hartungvideo1998,noorkamih264wm2007,tewh264hiding2014,duttahevcwm2016}, which buys robustness at the price of binding the scheme to that standard and of re-embedding whenever the video is transcoded. COVER keeps the flexibility of a learned watermark and resolves the tension by pairing a differentiable surrogate bank with latent re-encoding on a spatiotemporal generative prior~\citep{cogvideox2024,latentdiffusion2022}, so training targets the mechanisms realistic compression shares rather than one distortion standing in for every codec at once.

\section{Problem Formulation}
\label{sec:problem}

Given a video $\mathbf{x}\in\mathbb{R}^{T\times C\times H\times W}$ and a binary payload $\mathbf{b}\in\{0,1\}^{L}$, our goal is a marked video $\mathbf{x}_{w}$ that stays visually close to $\mathbf{x}$ and from which $\mathbf{b}$ can still be recovered after real codec compression, where $T$, $C$, $H$, and $W$ are the frame, channel, height, and width counts and $L$ is the payload length. A realistic distribution pipeline degrades the mark in two stages. Rendering a marked latent into the pixel domain and reading it back introduces a reconstruction gap, and lossy compression then removes spatial and temporal detail through quantization and the other steps of a codec. A decoder trained on the clean latent alone meets neither stage, which is the failure the three recovery paths below are built to prevent.

We consider a provider that marks its own videos and later verifies them, as in provenance and attribution for generated media. The provider holds the frozen encoder and the trained latent decoder, so detection is informed and needs no access to the source video, whereas the codec in between is a black box whose standard and strength are unknown and whose steps carry no gradient. We model the distribution channel rather than an adversary who sets out to erase the mark, so removal attacks and geometric editing both lie outside the scope of what this present paper sets out to study, and we do not evaluate them here.

\section{Method}
\label{sec:method}

\subsection{Overview}

COVER addresses both stages through three shared recovery paths, as illustrated in Figure~\ref{fig:framework}. The latent path checks that the payload is present in the marked latent. The re-encoding path sends the marked video through the frozen video decoder and encoder, so the latent decoder learns to tolerate the reconstruction gap. The codec path additionally applies a differentiable surrogate in the pixel domain before re-encoding, so the latent decoder learns to tolerate compression. We keep the pretrained encoder and decoder frozen and train only the lightweight modules on top, namely a payload encoder and an adapter that together form the embedder, one latent decoder shared by the three paths, and two discriminators. Freezing the autoencoder preserves its generative prior and confines watermark learning to compact modules, which keeps training stable while the prior carries most of the representation.

\subsection{Latent Watermark Embedding}

Let $V_{\mathrm{enc}}$ and $V_{\mathrm{dec}}$ denote the frozen video encoder and decoder. The input video is mapped to a spatiotemporal latent $\mathbf{z}=V_{\mathrm{enc}}(\mathbf{x})\in\mathbb{R}^{T_z\times C_z\times H_z\times W_z}$, which is compressed relative to the input along time and space alike. A payload encoder maps the payload to a conditioning tensor $\mathbf{e}_{b}=E_{b}(\mathbf{b})$, broadcast to the spatiotemporal layout the adapter expects. The adapter reads the video latent together with the payload condition and predicts a payload dependent residual $\Delta\mathbf{z}_{w}=G_{\theta}(\mathbf{z},\mathbf{e}_{b})$, which the embedding step adds back to the video latent that the encoder produced,
\begin{equation}
    \mathbf{z}_{w}=\mathbf{z}+\alpha\,\Delta\mathbf{z}_{w},
    \label{eq:latent_embedding}
\end{equation}
where the strength $\alpha$ scales the residual. Residual embedding keeps the semantic structure of the pretrained latent while letting the adapter write payload dependent information into its stable components. The frozen decoder renders the marked video $\mathbf{x}_{w}=V_{\mathrm{dec}}(\mathbf{z}_{w})$ and the clean reconstruction $\mathbf{x}_{c}=V_{\mathrm{dec}}(\mathbf{z})$. We use $\mathbf{x}_{c}$ rather than the source $\mathbf{x}$ as the fidelity reference during training, which separates the perturbation caused by embedding from the reconstruction error that the frozen autoencoder introduces before any payload has been written into the latent at all.

\subsection{Robust Detection via Latent Re-Encoding}

The detector reads latent representations rather than pixel frames, so it reasons over the same spatiotemporal features used during embedding. It holds the frozen encoder used at embedding, which makes detection informed and independent of the source video. The re-encoding path maps the marked video back to the latent as $\hat{\mathbf{z}}_{w}=V_{\mathrm{enc}}(\mathbf{x}_{w})$, which exposes the decoder to the full decode then encode cycle and stops it from relying on latent detail that disappears once $\mathbf{z}_{w}$ becomes a pixel video. The codec path first passes the marked video through a differentiable surrogate $\mathcal{A}\sim\mathcal{S}$, drawn from the codec surrogate distribution $\mathcal{S}$ defined below, and then re-encodes it, giving $\tilde{\mathbf{z}}_{w}=V_{\mathrm{enc}}(\mathcal{A}(\mathbf{x}_{w}))$. A shared latent decoder $D_{\psi}$ reads each of these three latents in turn and produces one prediction per path,
\begin{equation}
    \hat{\mathbf{b}}_{\mathrm{lat}}=D_{\psi}(\mathbf{z}_{w}),\quad
    \hat{\mathbf{b}}_{\mathrm{re}}=D_{\psi}(\hat{\mathbf{z}}_{w}),\quad
    \hat{\mathbf{b}}_{\mathrm{cod}}=D_{\psi}(\tilde{\mathbf{z}}_{w}).
    \label{eq:predictions}
\end{equation}
The latent prediction supplies a direct capacity signal, while the re-encoding and codec predictions force the payload to survive the autoencoder and compression. Sharing $D_{\psi}$ encourages a single payload representation that stays readable in all three conditions rather than three specialized ones.

\subsection{Differentiable Codec Surrogate Bank}

\begin{figure}[t]
    \centering
    \includegraphics[width=0.98\columnwidth]{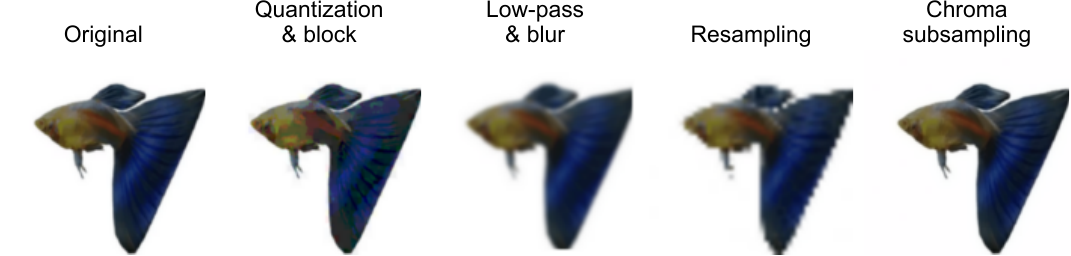}
    \caption{Effect of each codec surrogate group on an example frame. The strengths are exaggerated for visibility.}
    \label{fig:surrogate}
\end{figure}

Real codecs rely on discrete, non-differentiable operations, so they cannot be placed inside gradient-based training. Rather than reproduce any single codec, we build a differentiable surrogate bank that models the degradation mechanisms shared by practical compression, following the differentiable distortion layers used for image compression~\citep{distortionagnostic2020,mbrs2021}. Let $\mathcal{G}=\{g_{\mathrm{qb}},g_{\mathrm{lp}},g_{\mathrm{rs}},g_{\mathrm{ch}}\}$ denote four surrogate groups, and let $\mathcal{A}_{g}$ denote the differentiable operator that group $g$ applies at a sampled strength. A surrogate $\mathcal{A}$ chains $K\ge1$ such operators, with repetition allowed,
\begin{equation}
    \mathcal{A}=\mathcal{A}_{g_{K}}\circ\mathcal{A}_{g_{K-1}}\circ\cdots\circ\mathcal{A}_{g_{1}},\qquad g_{k}\in\mathcal{G}.
    \label{eq:surrogate_composition}
\end{equation}
The quantization and block group $g_{\mathrm{qb}}$ reduces value precision and adds locally structured block artifacts, which approximates the loss of fine coefficients and the discontinuities of block-based processing. The low pass group $g_{\mathrm{lp}}$ applies differentiable blur, which models the suppression of fine spatial detail at stronger compression. The resampling group $g_{\mathrm{rs}}$ performs differentiable downsampling and upsampling, which represents resolution adaptation during transcoding. The chroma group $g_{\mathrm{ch}}$ converts the video to a luminance and chrominance space, reduces chroma resolution or precision, and converts back, which approximates the weaker preservation of color under codecs that allocate fewer samples to chroma. Figure~\ref{fig:surrogate} shows the effect of each group on a frame.

The bank stores composite recipes, each chaining several operators at a preset strength, and the distribution $\mathcal{S}$ draws one recipe per training step from a fixed mixture that favors the harder compositions, so the latent decoder meets compound rather than isolated distortions in every forward pass. The harder recipes also add a small amount of noise, which stands for the residual coding error a codec leaves behind at a low bit rate. Every operator acts within a frame, so the bank models the per-frame degradation a codec makes visible and leaves inter-frame prediction to the temporal terms of the objective. These surrogates remain approximations that capture effects associated with H.264, H.265, AV1, and VP9 without reproducing the decisions any specific codec makes, and our evaluation therefore reports real codecs only.

\subsection{Training Objectives}

The full objective combines payload recovery with fidelity terms that regulate the residual in the pixel and frequency domains. Payload recovery drives the three shared paths through a binary cross entropy term applied to each of them in turn,
\begin{equation}
\begin{aligned}
    \mathcal{L}_{\mathrm{wm}}={}&\lambda_{\mathrm{lat}}\,\mathrm{BCE}(\mathbf{b},\hat{\mathbf{b}}_{\mathrm{lat}})+\lambda_{\mathrm{re}}\,\mathrm{BCE}(\mathbf{b},\hat{\mathbf{b}}_{\mathrm{re}})\\
    &+\lambda_{\mathrm{cod}}\,\mathrm{BCE}(\mathbf{b},\hat{\mathbf{b}}_{\mathrm{cod}}),
\end{aligned}
    \label{eq:watermark_loss}
\end{equation}
where $\lambda_{\mathrm{lat}}$, $\lambda_{\mathrm{re}}$, and $\lambda_{\mathrm{cod}}$ weight the latent, re-encoding, and codec paths, and are set so that the two re-encoded paths dominate the recovery term throughout the training.

\paragraph{Fidelity Terms.}
Let $\mathbf{r}=\mathbf{x}_{w}-\mathbf{x}_{c}$ denote the pixel residual introduced by embedding, and let every fidelity term be averaged over the mini-batch. A pixel term and a structural term bound its magnitude and protect local structure,
\begin{equation}
    \mathcal{L}_{\mathrm{mse}}=\frac{1}{TCHW}\|\mathbf{r}\|_{2}^{2},\qquad \mathcal{L}_{\mathrm{ssim}}=1-\mathrm{SSIM}(\mathbf{x}_{w},\mathbf{x}_{c}),
    \label{eq:quality_loss}
\end{equation}
where $\|\cdot\|_{2}$ is the Euclidean norm, $TCHW$ counts the pixel entries of a clip, and $\mathrm{SSIM}$ is the structural similarity index evaluated per frame and channel and averaged over the clip. A small frame residual can still flicker, so we also match the temporal differences of the two videos through the same residual. Writing $(\Delta\mathbf{r})_{t}=\mathbf{r}_{t+1}-\mathbf{r}_{t}$ for its first-order difference at frame $t$ and $(\Delta^{2}\mathbf{r})_{t}=\mathbf{r}_{t+2}-2\mathbf{r}_{t+1}+\mathbf{r}_{t}$ for its second-order difference, the two temporal terms are
\begin{equation}
\begin{aligned}
    \mathcal{L}_{\mathrm{tmp}}^{(1)}&=\frac{1}{T-1}\sum_{t=1}^{T-1}\|(\Delta\mathbf{r})_{t}\|_{1},\\
    \mathcal{L}_{\mathrm{tmp}}^{(2)}&=\frac{1}{T-2}\sum_{t=1}^{T-2}\|(\Delta^{2}\mathbf{r})_{t}\|_{1}.
\end{aligned}
    \label{eq:temporal_loss}
\end{equation}
The term $\mathcal{L}_{\mathrm{tmp}}^{(1)}$ aligns the frame-to-frame velocity of the two videos and $\mathcal{L}_{\mathrm{tmp}}^{(2)}$ aligns their acceleration, both under an $\ell_{1}$ norm over the valid frame pairs, so the marked video keeps the motion of the clean reconstruction and suppresses the temporal flicker that a spatial term alone would miss.

\paragraph{Spectral Shaping.}
High frequency watermark energy is the first to be removed by quantization and low pass degradation, so we penalize the residual spectrum by its radial frequency. Let $\hat{\mathbf{r}}_{t,c}=\mathcal{F}_{2\mathrm{D}}(\mathbf{r}_{t,c})$ denote the spatial Fourier transform of the residual at frame $t$ and channel $c$, let $\Omega$ be the set of frequency positions on the $H\times W$ grid, and $|\Omega|$ its cardinality. Each position $\omega\in\Omega$ carries a radial frequency $\rho_{\omega}\in[0,1]$ normalized by the Nyquist radius, and its weight $w(\rho_{\omega})=w_{\mathrm{low}}+(w_{\mathrm{high}}-w_{\mathrm{low}})\rho_{\omega}^{p}$ rises from $w_{\mathrm{low}}$ at low frequency to $w_{\mathrm{high}}$ at high frequency, with the exponent $p>0$ setting how sharply the penalty grows. The spectral term is
\begin{equation}
    \mathcal{L}_{\mathrm{freq}}=\frac{1}{TC|\Omega|}\sum_{t,c}\sum_{\omega\in\Omega}w(\rho_{\omega})\,|\hat{\mathbf{r}}_{t,c}(\omega)|^{2},
    \label{eq:frequency_loss}
\end{equation}
which steers watermark energy toward the low and mid frequencies that a codec is built to preserve in any case.

\paragraph{Adversarial Realism.}
A frame patch discriminator $D_{f}$ scores single frames and a spatiotemporal discriminator $D_{v}$ scores the whole clip, both taking the clean reconstruction as the real sample. The generator minimizes $\mathcal{L}_{\mathrm{adv}}^{f}=-\frac{1}{T}\sum_{t=1}^{T}D_{f}(\mathbf{x}_{w,t})$ and $\mathcal{L}_{\mathrm{adv}}^{v}=-D_{v}(\mathbf{x}_{w})$, and each discriminator $D\in\{D_{f},D_{v}\}$ is trained with a hinge loss that pushes the score of the clean reconstruction above the score of the marked video by a unit margin,
\begin{equation}
    \mathcal{L}_{D}=\mathbb{E}\!\left[\max(0,1-D(\mathbf{x}_{c}))\right]+\mathbb{E}\!\left[\max(0,1+D(\mathbf{x}_{w}))\right].
    \label{eq:disc_loss}
\end{equation}

\paragraph{Overall Objective.}
The trainable modules minimize a weighted sum of the recovery and fidelity terms,
\begin{equation}
\begin{aligned}
    \mathcal{L}_{\mathrm{total}}={}&\mathcal{L}_{\mathrm{wm}}+\lambda_{\mathrm{mse}}\mathcal{L}_{\mathrm{mse}}+\lambda_{\mathrm{ssim}}\mathcal{L}_{\mathrm{ssim}}\\
    &+\lambda_{\mathrm{tmp}}^{(1)}\mathcal{L}_{\mathrm{tmp}}^{(1)}+\lambda_{\mathrm{tmp}}^{(2)}\mathcal{L}_{\mathrm{tmp}}^{(2)}+\lambda_{\mathrm{freq}}\mathcal{L}_{\mathrm{freq}}\\
    &+\lambda_{\mathrm{adv}}^{f}\mathcal{L}_{\mathrm{adv}}^{f}+\lambda_{\mathrm{adv}}^{v}\mathcal{L}_{\mathrm{adv}}^{v},
\end{aligned}
    \label{eq:total_loss}
\end{equation}
where each coefficient sets its influence and the implementation details list the values. Algorithm~\ref{alg:training} in the appendix summarizes the procedure, in which the generator and the two discriminators alternate while the frozen autoencoder receives no gradients, so its generative prior is unchanged from the first iteration of training through to the very last one.

\section{Experiments}
\label{sec:experiments}

\subsection{Experimental Setup}

\paragraph{Benchmark.}
We evaluate on 500 videos, each processed by 12 real codec settings drawn from four codec families. The videos are rendered by a text-to-video model from a fixed prompt set at $256\times256$ with 16 frames per clip, and the training clips come from a disjoint prompt set, so no test video is seen during training. We drive each codec through its rate control parameter, namely the constant rate factor (CRF) for AV1, H.264, and H.265, and the constant quality (CQ) level for VP9. We use CRF values of 45, 55, and 63 for AV1, 23, 35, and 45 for H.264, and 26, 40, and 50 for H.265, together with CQ values of 35, 45, and 55 for VP9. A larger CRF or CQ denotes stronger compression within a family and is not comparable across families. Each video and codec pair is tested with 100 random payloads, giving $500\times12\times100=600{,}000$ payload trials and $4{,}800{,}000$ evaluated bits, so the numbers that we report for each of the 12 settings rest on 50,000 payload trials and 400,000 evaluated bits drawn for that setting alone.

\paragraph{Models.}
We build COVER on the frozen video autoencoder of CogVideoX-2b~\citep{cogvideox2024}, whose encoder divides the frame count by four and each spatial dimension by eight and returns 16 latent channels, so a 16-frame clip at $256\times256$ becomes a latent on a $32\times32$ grid. On top of this prior we train a small set of watermarking modules. The payload encoder maps the payload to a conditioning tensor through a learned codebook, the adapter is a small convolutional network that reads the concatenation of the latent and that tensor and predicts the residual, and the latent decoder is a spatiotemporal network pooling the re-encoded latent into payload logits. The frame discriminator is a patch network over single frames, and the spatiotemporal discriminator is a three-dimensional network judging coherence across the clip.

\paragraph{Baselines.}
We compare COVER against four recent video watermarking systems, namely VideoMark~\citep{videomark2025}, VideoShield~\citep{videoshield2025}, Video Seal~\citep{videoseal2024}, and VideoSignature~\citep{videosignature2025}. VideoMark, VideoShield, and VideoSignature place the watermark inside video generation, so they can mark only the clips they synthesize, and we let them regenerate the benchmark clips from the prompts that produced our test videos. Video Seal marks a video that already exists, which is also the setting of COVER, so both read the rendered clips directly. All four meet the codec protocol described above, so every method is stressed in exactly the same way.

\paragraph{Metrics.}
We use bit accuracy as the main robustness metric. With $N$ videos and $M$ payloads per video and codec pair, each payload of length $L$, we define bit accuracy as
\begin{equation}
\mathrm{BA}=\frac{1}{NML}\sum_{i=1}^{N}\sum_{m=1}^{M}\sum_{j=1}^{L}\mathds{1}\!\left(\hat{b}_{i,m,j}=b_{i,m,j}\right),
\label{eq:bit_accuracy}
\end{equation}
where $\mathds{1}(\cdot)$ equals one when the predicted bit is correct, and our protocol sets $N=500$, $M=100$, and $L=8$. We also report the average and the worst bit accuracy over the 12 settings, and we measure visual quality with PSNR in dB and SSIM against the source video before compression. A verifier turns bit accuracy into a decision by comparing the extracted payload with a registered one, and under an exact rule that accepts only when all $L$ bits agree an unmarked video passes with probability $2^{-L}$, or 0.39\% at $L=8$, and relaxing it to at least $L-1$ agreeing bits raises that probability to $(L+1)2^{-L}$, or 3.52\%. We also report the proportion of marked videos a verifier accepts under this rule as detection accuracy. Bit accuracy says how much of the payload survives and detection accuracy says how often the whole decision comes out right, so a short payload can hold a high bit accuracy and still lose the decision that a verifier actually has to take.

\paragraph{Implementation Details.}
Following prior work~\citep{videoshield2025,videosignature2025}, each clip contains 16 frames. The payload length is $L=8$ bits and the residual strength is $\alpha=0.10$. Optimization uses AdamW at a learning rate of $1\times10^{-4}$ with an effective batch size of eight, a fixed seed of 1234, and 5,000 steps. The recovery weights are $\lambda_{\mathrm{lat}}=1.0$ and $\lambda_{\mathrm{re}}=\lambda_{\mathrm{cod}}=3.0$, so the two re-encoding paths outweigh the direct one. The fidelity weights are $\lambda_{\mathrm{mse}}=0.0015$, $\lambda_{\mathrm{ssim}}=0.015$, $\lambda_{\mathrm{tmp}}^{(1)}=0.01$, $\lambda_{\mathrm{tmp}}^{(2)}=0.005$, and $\lambda_{\mathrm{freq}}=0.001$, both adversarial weights are 0.0005, and the spectral term uses $w_{\mathrm{high}}=1.0$, $w_{\mathrm{low}}=0.1$, and $p=2.0$. All experiments run on NVIDIA RTX PRO Blackwell GPUs.

\begin{table*}[t]
\caption{Bit accuracy (\%) under 12 real codec settings from four codec families. Bold and underline mark the accuracy columns only. Training never uses a real codec, so every setting here tests how far the surrogate bank transfers to an unseen encoder.}
\label{tab:main_results}
\centering
\footnotesize
\resizebox{0.98\textwidth}{!}{%
\setlength{\tabcolsep}{3pt}
\begin{tabular}{lcccccccccccccccc}
\toprule
\multirow{2}{*}{\textbf{Method}} &
\multicolumn{3}{c}{\textbf{AV1 (CRF)}} &
\multicolumn{3}{c}{\textbf{H.264 (CRF)}} &
\multicolumn{3}{c}{\textbf{H.265 (CRF)}} &
\multicolumn{3}{c}{\textbf{VP9 (CQ)}} &
\multirow{2}{*}{\textbf{Average}} &
\multirow{2}{*}{\textbf{Worst}} &
\multirow{2}{*}{\textbf{PSNR}} &
\multirow{2}{*}{\textbf{SSIM}} \\
\cmidrule(lr){2-4}\cmidrule(lr){5-7}\cmidrule(lr){8-10}\cmidrule(lr){11-13}
& \textbf{45} & \textbf{55} & \textbf{63} & \textbf{23} & \textbf{35} & \textbf{45} & \textbf{26} & \textbf{40} & \textbf{50} & \textbf{35} & \textbf{45} & \textbf{55} & & & & \\
\midrule
VideoMark
& 17.40 & 8.10 & 1.80
& 56.90 & 10.30 & 0.40
& 42.30 & 3.90 & 0.60
& 38.30 & 19.50 & 6.30
& 17.15 & 0.40 & 34.27 & 0.866 \\
VideoShield
& \underline{97.80} & \underline{92.90} & \underline{77.80}
& \textbf{99.80} & \underline{96.80} & \underline{75.90}
& \underline{99.40} & \underline{91.60} & \underline{68.90}
& \underline{99.40} & \underline{98.20} & \underline{93.90}
& \underline{91.03} & \underline{68.90} & 34.03 & 0.860 \\
Video Seal
& 67.50 & 60.00 & 51.30
& 94.20 & 62.10 & 50.50
& 86.40 & 55.20 & 50.20
& 88.70 & 77.00 & 62.10
& 67.10 & 50.20 & 30.55 & 0.812 \\
VideoSignature
& 50.90 & 47.20 & 44.60
& 59.30 & 49.70 & 44.10
& 57.40 & 46.80 & 45.90
& 58.70 & 53.10 & 49.00
& 50.56 & 44.10 & 25.02 & 0.808 \\
\midrule
COVER (Ours)
& \textbf{98.91} & \textbf{97.24} & \textbf{83.23}
& \underline{99.74} & \textbf{97.99} & \textbf{85.29}
& \textbf{99.67} & \textbf{96.41} & \textbf{73.72}
& \textbf{99.50} & \textbf{98.35} & \textbf{94.55}
& \textbf{93.72} & \textbf{73.72} & 28.30 & 0.855 \\
\bottomrule
\end{tabular}
}
\end{table*}

\subsection{Main Results}

Table~\ref{tab:main_results} reports bit accuracy under all 12 real codec settings, together with the aggregate accuracy and the visual quality. COVER attains the highest average bit accuracy of 93.72\%, equivalently an average bit error rate of 6.28\%, and it ranks first on 11 of the 12 settings. The only exception is H.264 at CRF 23, where COVER and the best method are effectively tied above 99.7\%. COVER improves the strongest prior method, VideoShield, by 2.68 points in average bit accuracy and raises the worst case from 68.90\% to 73.72\%. Bit accuracy decreases as compression strengthens within each family, as expected, yet COVER stays above 90\% on nine settings and above 80\% on 11 of them. The marked videos keep a PSNR of 28.30 and an SSIM of 0.855 before compression, and Figure~\ref{fig:quality} shows that a marked frame stays visually close to its source. Because the source is the reference here, the reported PSNR also absorbs the reconstruction error of the frozen autoencoder, a fidelity floor that is already in place before any payload bit is written at all.

The margin is consistent rather than driven by a single setting. Averaged within each family, COVER reaches 93.13\% on AV1, 94.34\% on H.264, 89.93\% on H.265, and 97.47\% on VP9, improving VideoShield in every one of them. How large it becomes depends on how hard the setting is. At the mild settings both methods sit above 97\% and separate by a few tenths of a point, so the comparison there says little, whereas the gap opens to 9.39 points at H.264 CRF 45 and to 5.43 points at AV1 CRF 63, the regime that decides whether a watermark survives once a video is stored and shared.

AV1 and H.265 at their strongest settings stress the watermark most, since aggressive transform quantization removes the fine detail a pixel domain watermark would depend on. COVER still keeps 83.23\% on AV1 CRF 63 and 73.72\% on H.265 CRF 50, its lowest result, because the payload lives in the latent representation the frozen prior reconstructs rather than in the discarded detail. VP9 is the easiest family in our range, where COVER stays above 94\% throughout. VideoMark performs worst by a wide margin, since its payload is carried by the initial generation noise and read back by temporal matching, and neither survives the range we test. Its accuracy falls below the 50\% a detector should approach once the payload is gone, so we report the row as measured and read it as a floor on the method rather than a characterization of it. Video Seal falls toward 50\% wherever compression is strong and VideoSignature stays near chance, which shows that robustness to real compression across several codecs does not follow from image style watermarking on its own.

Detection accuracy reads differently, and Table~\ref{tab:detection_accuracy} in the appendix gives it for every setting. COVER reaches 71.13\% on average and comes second to VideoShield at 96.92\%, ahead of Video Seal by 15.96 points. The gap comes from the payload length rather than from how well the mark survives, since a verifier that has to read all eight bits fails as soon as one of them is wrong, and the appendix works this arithmetic out in detail.

The strongest setting of each family gives a compact reading of the worst realistic case, which is the compression a video meets after transcoding for storage or streaming. Averaged over those four settings COVER reaches 84.20\% bit accuracy against 79.13\% for VideoShield, and it holds a worst case of 73.72\% where the strongest prior method drops to 68.90\%. A higher floor matters more than a higher peak, since a watermark is only as reliable as its weakest operating point, and the gap widens exactly where bit accuracy is under the most pressure, which is what a provider needs when it controls neither the codec nor the quality setting a downstream platform applies. No real codec ever enters training, yet the ranking holds across all four standards, which indicates that the surrogate captures degradation shared by modern compression rather than the quirks of one implementation.

\subsection{Visual Quality}

\begin{figure}[t]
    \centering    \includegraphics[width=0.98\columnwidth]{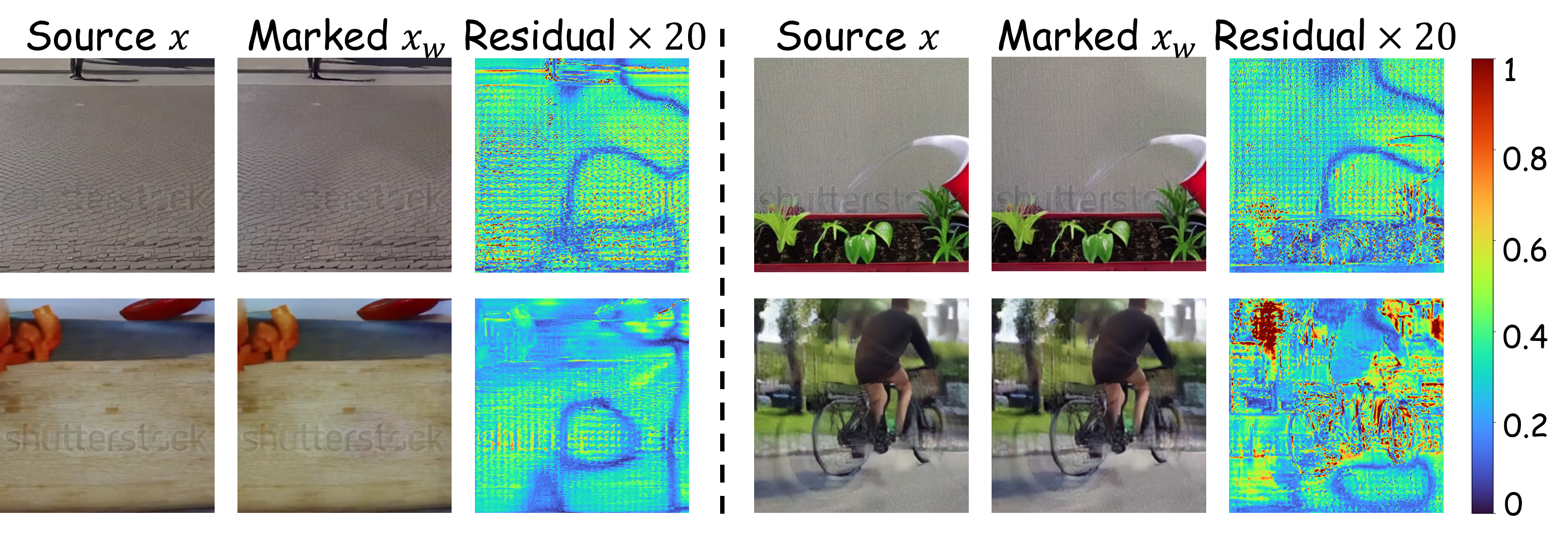}
    \caption{Qualitative comparison for four clips. Each triplet shows the source frame, the marked frame, and their residual, magnified 20 times and rendered under a color map.}
    \label{fig:quality}
\end{figure}

Figure~\ref{fig:quality} compares source frames, marked frames, and the residual between them for four clips. A marked frame is hard to separate from its source at normal viewing, and the residual becomes visible only under magnification and a color map. Its amplitude peaks along edges and fine detail, such as the leaves and the rim of the planter, and drops on smooth bright surfaces, so the watermark sits where a viewer is least equipped to notice it, while away from those contours it settles into a fine regular texture consistent with a payload written on the grid of the latent rather than on pixels. Figure~\ref{fig:more_quality} in the appendix follows six further clips over eight consecutive frames, where the residual keeps its character from frame to frame. The faint text across some frames is a stock footage watermark that the text-to-video model reproduces from its training data, and it appears in the source and the marked frame alike, so it belongs to the content rather than to anything COVER writes. The remaining gap to the source comes mostly from the frozen autoencoder, which already reconstructs a video with a small error of its own before any watermark is added to it at all.

\subsection{Ablation Study}
\label{sec:ablation}

Each variant changes only the named component, with all other settings fixed and the same pretrained initialization.

\paragraph{Surrogate Components.}
We disable one codec surrogate group at a time and report the result of retraining in Table~\ref{tab:surrogate_ablation}. Every group contributes, and the quantization and block group carries the most weight by a wide margin, since removing it costs 4.66 points on the average and 8.40 points on the worst case. This is what the design predicts, since transform quantization is applied first and hardest. The low pass and the resampling groups follow at 2.15 and 1.73 points. Chroma is the only group that does not help everywhere, since removing it costs 0.82 points on the average yet raises the worst case to 74.66\%. That setting is H.265 at CRF 50, where the damage comes mostly from luma quantization, so a bank that also trains against chroma reduction spends part of its capacity on a mechanism that this particular setting never stresses.

\begin{table}[t]
\caption{Ablation of the codec surrogate groups, with one group disabled per row. Bit accuracy in \%, best in bold.}
\label{tab:surrogate_ablation}
\centering
\small
\setlength{\tabcolsep}{4pt}
\begin{tabular}{lcc}
\toprule
\textbf{Surrogate bank} & \textbf{Average} & \textbf{Worst} \\
\midrule
COVER (full)                   & \textbf{93.72} & 73.72 \\
Without quantization and block & 89.06 & 65.32 \\
Without low pass and blur      & 91.57 & 70.30 \\
Without resampling             & 91.99 & 71.28 \\
Without chroma                 & 92.90 & \textbf{74.66} \\
\bottomrule
\end{tabular}
\end{table}

\paragraph{Loss Families.}
We remove one loss family at a time and report the effect in Table~\ref{tab:loss_ablation}. The spectral term matters most for robustness, since dropping it costs 2.57 points on the average and 5.01 points on the worst case, which is our strongest evidence that steering the payload away from the high frequencies is what carries it through a codec. The other three cost between 1.09 and 1.56 points. Fidelity behaves differently, as every variant gives up about 3.5 dB of PSNR and the four land within 0.05 dB of one another, so the fidelity terms act together and no single one accounts for the marked video's quality. The temporal errors separate them far less, since T-E1 spans 7.36 to 7.91 across all five, and the variant without adversarial terms is lowest on both. Those terms push the residual toward frame-level realism and let it vary from frame to frame, so dropping them leaves a more static residual, smoother in time though its PSNR is 3.52 dB worse.

\begin{table}[t]
\caption{Ablation of the loss families, one per row. Bit accuracy in \%. T-E1 and T-E2 are the first- and second-order differences of the residual in $10^{-3}$, where lower is better and the best value in each column is in bold.}
\label{tab:loss_ablation}
\centering
\small
\setlength{\tabcolsep}{3pt}
\begin{tabular}{lccccc}
\toprule
\textbf{Objective} & \textbf{Average} & \textbf{Worst} & \textbf{PSNR} & \textbf{T-E1} & \textbf{T-E2} \\
\midrule
COVER (full)        & \textbf{93.72} & \textbf{73.72} & \textbf{28.30} & 7.83 & 11.66 \\
Without spatial     & 92.63 & 71.98 & 24.80 & 7.39 & 11.26 \\
Without temporal    & 92.16 & 71.23 & 24.77 & 7.82 & 11.59 \\
Without spectral    & 91.15 & 68.71 & 24.82 & 7.91 & 11.69 \\
Without adversarial & 92.60 & 71.43 & 24.78 & \textbf{7.36} & \textbf{11.23} \\
\bottomrule
\end{tabular}
\end{table}

\section{Discussion}
\label{sec:discussion}

\paragraph{Why the Design Works.}
The results support the introduction's intuition. The advantage is largest at the strongest setting of each family, where aggressive quantization strips the detail a frequency domain watermark relies on, while the latent representation and the spectral term hold the payload inside the band compression preserves, which the loss ablation confirms, since removing no other loss family costs as much robustness as removing that single spectral term from the objective.

\paragraph{Surrogate and Real Codecs.}
The bank never reproduces a codec, and the distance between them is where this design could have failed. Every surrogate operator acts inside a frame, whereas a real encoder also spends its budget across frames through motion compensation, so training sees a narrower channel than the test one does. That the ranking still holds on four standards suggests the payload depends on no single frame, since it is carried by latent components the frozen prior reconstructs from surviving structure, while the temporal terms hold the marked clip to the motion of the clean one, so a predictor coding a frame from its neighbors meets a residual field that moves along with the content itself.

\paragraph{Practical Scope.}
COVER marks a video that already exists, which the strongest baseline cannot do, since VideoShield plants its payload in the noise a generator starts from and can therefore mark only what that generator produces. Embedding costs one encoder pass and one decoder pass of the frozen autoencoder, and verification costs one encoder pass followed by the latent decoder, so a provider checks a received clip in a single forward sweep. The detector is tied to the autoencoder used at embedding, so moving to another video prior asks for a short retraining of the watermarking modules on whatever latent space that new prior happens to define for them to write the payload into.

\paragraph{Limitations and Future Work.}
COVER carries eight bits, which is shorter than the payload the baselines embed, so Table~\ref{tab:main_results} compares robustness at each system's operating capacity rather than at a matched one, and longer payloads would show how gracefully accuracy degrades as capacity grows. Its marked videos also sit below the two strongest baselines on PSNR, so the comparison trades fidelity for robustness, and a matched quality protocol that sweeps the residual strength would separate the two axes. We also report bit accuracy rather than a decision rate, since eight bits leave a per-video verdict too thin to calibrate, and a longer payload is the natural route to a deployable detector. The evaluation itself targets compression at one resolution and clip length, leaving longer clips, higher resolutions, everyday geometric edits, and a determined removal adversary all still open.

\section{Conclusion}
\label{sec:conclusion}

We presented COVER, a video watermark whose design target is codec compression. Its payload is written into the latent of a frozen generative video autoencoder and recovered by re-encoding the received video into that space, under three shared recovery paths whose codec branch draws on a surrogate bank. Across four codecs at 12 settings COVER reaches 93.72\% average bit accuracy, ranks first on 11, and improves the strongest prior method by 2.68 points, while the ablations name the quantization surrogate and the spectral term as the two components that matter the most here. Two extensions follow directly from this position. A longer payload would turn the accuracies reported here into calibrated per-video decisions, and a surrogate bank that also models inter-frame prediction would close the one dimension in which our training channel stays narrower than a real encoder, so that a codec-first watermark becomes not merely a robustness result but a provenance tool that a real platform can rely on.

\bibliographystyle{ACM-Reference-Format}
\bibliography{refs}

\clearpage

\appendix
\section{Appendix}

\subsection{Training Procedure}
\label{app:algorithm}

Algorithm~\ref{alg:training} states the training loop referenced in the method section. The generator modules and the two discriminators are updated in alternation on every iteration, and the frozen autoencoder receives no gradients at any point, so the generative prior it supplies is identical before and after training.

\begin{algorithm}[h]
\caption{COVER Training}
\label{alg:training}
\begin{algorithmic}[1]
\REQUIRE Training set $\mathcal{D}$; iterations $N_{\mathrm{it}}$; payload length $L$; frozen $V_{\mathrm{enc}},V_{\mathrm{dec}}$; trainable $G_{\theta},E_{b},D_{\psi}$; surrogate distribution $\mathcal{S}$; discriminators $D_{f},D_{v}$; strength $\alpha$
\ENSURE Trained $G_{\theta},E_{b},D_{\psi}$
\FOR{$i=1$ \TO $N_{\mathrm{it}}$}
    \STATE Sample video batch $\mathbf{x}\sim\mathcal{D}$ and payload $\mathbf{b}\sim\mathrm{Bernoulli}(0.5)^{L}$
    \STATE $\mathbf{z}\leftarrow V_{\mathrm{enc}}(\mathbf{x})$,\quad $\mathbf{z}_{w}\leftarrow \mathbf{z}+\alpha\,G_{\theta}(\mathbf{z},E_{b}(\mathbf{b}))$
    \STATE $\mathbf{x}_{c}\leftarrow V_{\mathrm{dec}}(\mathbf{z})$,\quad $\mathbf{x}_{w}\leftarrow V_{\mathrm{dec}}(\mathbf{z}_{w})$
    \STATE $\hat{\mathbf{z}}_{w}\leftarrow V_{\mathrm{enc}}(\mathbf{x}_{w})$
    \STATE $\mathcal{A}\sim\mathcal{S}$,\; $\tilde{\mathbf{z}}_{w}\leftarrow V_{\mathrm{enc}}(\mathcal{A}(\mathbf{x}_{w}))$
    \STATE $\hat{\mathbf{b}}_{\mathrm{lat}},\hat{\mathbf{b}}_{\mathrm{re}},\hat{\mathbf{b}}_{\mathrm{cod}}\leftarrow D_{\psi}(\mathbf{z}_{w}),D_{\psi}(\hat{\mathbf{z}}_{w}),D_{\psi}(\tilde{\mathbf{z}}_{w})$
    \STATE Compute $\mathcal{L}_{\mathrm{total}}$ from \\ \quad $\mathbf{b},\hat{\mathbf{b}}_{\mathrm{lat}},\hat{\mathbf{b}}_{\mathrm{re}},\hat{\mathbf{b}}_{\mathrm{cod}},\mathbf{x}_{w},\mathbf{x}_{c},D_{f},D_{v}$ by Eq.~\ref{eq:total_loss}
    \STATE Update $G_{\theta},E_{b},D_{\psi}$ by $\nabla\mathcal{L}_{\mathrm{total}}$
    \STATE Compute $\mathcal{L}_{D}$ on detached $\mathbf{x}_{c},\mathbf{x}_{w}$ by Eq.~\ref{eq:disc_loss}
    \STATE Update $D_{f},D_{v}$ by $\nabla\mathcal{L}_{D}$
\ENDFOR
\end{algorithmic}
\end{algorithm}

\begin{table*}[!t]
\caption{Detection accuracy (\%) under 12 real codec settings from four codec families. Best in bold, second best underlined.}
\label{tab:detection_accuracy}
\centering
\footnotesize
\resizebox{0.92\textwidth}{!}{%
\setlength{\tabcolsep}{3pt}
\begin{tabular}{lcccccccccccccc}
\toprule
\multirow{2}{*}{\textbf{Method}} &
\multicolumn{3}{c}{\textbf{AV1 (CRF)}} &
\multicolumn{3}{c}{\textbf{H.264 (CRF)}} &
\multicolumn{3}{c}{\textbf{H.265 (CRF)}} &
\multicolumn{3}{c}{\textbf{VP9 (CQ)}} &
\multirow{2}{*}{\textbf{Average}} &
\multirow{2}{*}{\textbf{Worst}} \\
\cmidrule(lr){2-4}\cmidrule(lr){5-7}\cmidrule(lr){8-10}\cmidrule(lr){11-13}
& \textbf{45} & \textbf{55} & \textbf{63} & \textbf{23} & \textbf{35} & \textbf{45} & \textbf{26} & \textbf{40} & \textbf{50} & \textbf{35} & \textbf{45} & \textbf{55} & & \\
\midrule
VideoMark
& 7.37 & 0.00 & 0.00
& 42.11 & 4.21 & 0.00
& 29.47 & 0.00 & 0.00
& 25.26 & 10.53 & 0.00
& 9.91 & 0.00 \\
VideoShield
& \textbf{100.00} & \textbf{99.00} & \textbf{90.00}
& \textbf{100.00} & \textbf{100.00} & \textbf{92.00}
& \textbf{100.00} & \textbf{100.00} & \textbf{82.00}
& \textbf{100.00} & \textbf{100.00} & \textbf{100.00}
& \textbf{96.92} & \textbf{82.00} \\
Video Seal
& 67.00 & 45.00 & 3.00
& \textbf{100.00} & 64.00 & 1.00
& \underline{99.00} & 26.00 & 2.00
& \underline{99.00} & \underline{93.00} & 63.00
& 55.17 & 1.00 \\
VideoSignature
& 0.00 & 0.00 & 0.00
& 2.00 & 0.00 & 0.00
& 2.00 & 0.00 & 0.00
& 5.00 & 0.00 & 0.00
& 0.75 & 0.00 \\
\midrule
COVER (Ours)
& \underline{92.00} & \underline{81.53} & \underline{27.43}
& \underline{98.00} & \underline{86.01} & \underline{32.13}
& 97.53 & \underline{76.24} & \underline{11.11}
& 96.28 & 88.10 & \underline{67.20}
& \underline{71.13} & \underline{11.11} \\
\bottomrule
\end{tabular}
}
\end{table*}

\subsection{Detection Accuracy}
\label{app:detection}

Table~\ref{tab:detection_accuracy} reports the proportion of marked videos a verifier accepts, over the same 12 settings as Table~\ref{tab:main_results}. COVER averages 71.13\% and is second to VideoShield at 96.92\%, while Video Seal reaches 55.17\% and the two remaining baselines stay near zero. COVER therefore leads every method except VideoShield, and it does so while holding the higher bit accuracy on 11 of the 12 settings, as Table~\ref{tab:main_results} shows.

The reason it still loses this column is the payload length, and the numbers show it directly. Our verifier accepts only when all eight bits are right, so detection accuracy is close to bit accuracy raised to the eighth power. On AV1 CRF 45 a bit accuracy of 98.91\% predicts 91.61\% and we measure 92.00\%, and on H.265 CRF 50 a bit accuracy of 73.72\% predicts 8.72\% and we measure 11.11\%. Averaged over the 12 settings the prediction is 69.48\% against 71.13\% measured, and the small excess in every setting shows that bit errors are mildly correlated. Detection therefore collapses where bit accuracy dips, which is why this column falls off at the strongest AV1 and H.265 settings while the bit accuracy column still holds up there.

VideoShield differs on this one point. It plants a much longer payload in the noise its generator starts from, so its verifier can decide on the weight of many bits instead of on every bit, and a bit accuracy of 91.03\% is still a safe decision when it is spread over hundreds of them. Eight bits leave no such room. Payload length is what separates the two methods here. COVER puts more of the payload through the codec intact and still loses the decision, because it has fewer bits to lose. VideoShield pays for its length in another way, since it marks only the clips it generates and cannot mark a video that already exists, which is precisely the setting that COVER is built to serve.

The detection column is where the next work should go. A longer payload with an error correcting code would let a verifier accept on a majority of bits, and the bit accuracy COVER already holds would then be enough to settle the decision at the hard settings too.

\begin{figure*}[t]
    \centering
    \vspace{2mm}
    \includegraphics[width=0.98\textwidth]{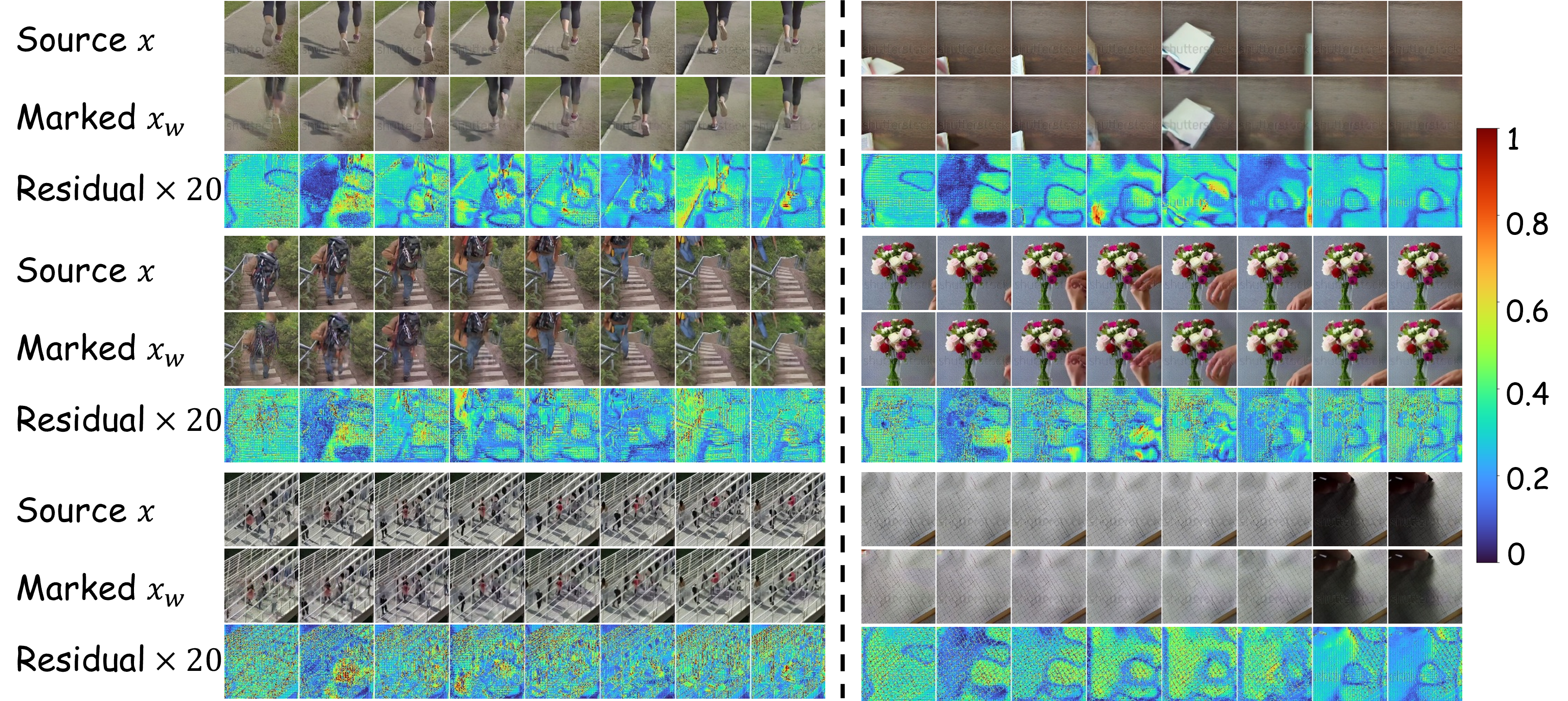}
    \caption{More examples of qualitative comparison. Each video clip is shown by eight frames.}
    \label{fig:more_quality}
\end{figure*}

\subsection{Additional Qualitative Results}
\label{app:more_examples}

Figure~\ref{fig:more_quality} extends the qualitative comparison to six further clips and follows each one over eight consecutive frames, so the residual can be judged over a whole clip instead of a single frame. The residual keeps the same fine lattice from one frame to the next instead of switching on and off between neighbours, which is the behaviour the two temporal terms are meant to produce. Its peaks travel with the content, sitting on the feet of the runner in the first clip and on the moving hand in the flower clip, while smooth regions such as the wall behind the flowers stay low throughout. How much residual a clip carries also depends on what it shows, since a frame full of texture such as the staircase behind the mesh takes more than a frame that is mostly flat, which is where a change would be easiest for a viewer to see.

\end{document}